\documentclass[11pt,a4paper]{article}
\usepackage[hyperref]{acl2018}
\usepackage{times}
\usepackage{latexsym}
\usepackage{graphicx}
\usepackage{url}
\usepackage{amsfonts}
\usepackage{multirow}
\usepackage{subfigure}
\usepackage{float}
\usepackage{caption}
\usepackage{amsmath}
\usepackage{makecell}
\usepackage{longtable}
\usepackage{verbatim}
\usepackage{bm}

\aclfinalcopy 
\def\aclpaperid{177} 

\title{Hybrid semi-Markov CRF for Neural Sequence Labeling}

\author{Zhi-Xiu Ye \\
  University of Science and \\
  Technology of China \\
  {\tt zxye@mail.ustc.edu.cn} \\\And
  Zhen-Hua Ling \\
  University of Science and \\
  Technology of China \\
  {\tt zhling@ustc.edu.cn} \\}

\date{}

\begin{document}
\maketitle
\begin{abstract}
  This paper proposes hybrid semi-Markov conditional random fields (SCRFs) for neural sequence labeling in natural language processing.
  Based on conventional conditional random fields (CRFs), SCRFs have been designed for the tasks of assigning labels to segments by extracting features from and describing transitions between segments instead of words.
  In this paper, we improve the existing SCRF methods by employing word-level and segment-level information simultaneously.
  First, word-level labels are utilized to derive the segment scores in SCRFs.
  Second, a CRF output layer and an SCRF output layer are integrated into an unified neural network and trained jointly.
  Experimental results on CoNLL 2003 named entity recognition (NER) shared task show that our model achieves state-of-the-art performance
  when no external knowledge is used\footnote{The code of our models is available at \url{http://github.com/ZhixiuYe/HSCRF-pytorch}}.
\end{abstract}

\section{Introduction}

Sequence labeling, such as part-of-speech (POS) tagging, chunking, and named entity recognition (NER), is a category of fundamental tasks in natural language processing (NLP).
Conditional random fields (CRFs) \cite{lafferty2001conditional}, as probabilistic undirected graphical models, have been widely applied to the sequence labeling tasks considering that they are able to describe the dependencies between adjacent word-level labels and to avoid illegal label combination
(e.g., I-ORG can't follow B-LOC in the NER tasks using the BIOES tagging scheme).
Original CRFs utilize hand-crafted features 
which increases the difficulty of performance tuning and domain adaptation.
In recent years, neural networks with distributed word representations  (i.e., word embeddings) \cite{mikolov2013distributed, pennington2014glove}
have been introduced to calculate word scores automatically for CRFs \cite{chiu2015named,huang2015bidirectional}.

On the other hand, semi-Markov conditional random fields (SCRFs) \cite{sarawagi2005semi} have been proposed for the tasks of assigning labels to the segments of input sequences,
e.g., NER.
Different from CRFs, SCRFs adopt segments instead of words as the basic units for feature extraction and transition modeling.
The word-level transitions within a segment are usually ignored.
Some variations of SCRFs have also been studied. For example,
\citet{andrew2006hybrid} extracted segment-level features by combining hand-crafted CRF features and modeled the Markov property between words instead of segments in SCRFs.
With the development of deep learning, some models of combining neural networks and SCRFs have also been studied.
\citet{zhuo2016segment} and \citet{kong2015segmental}  employed gated recursive convolutional neural networks (grConvs) and segmental recurrent neural networks (SRNNs) to  calculate segment scores for SCRFs respectively.

All these existing neural sequence labeling methods using SCRFs only adopted segment-level labels for score calculation and model training.
In this paper, we suppose that word-level labels can also contribute to the building of SCRFs and
thus design a hybrid SCRF (HSCRF) architecture for neural sequence labeling.
In an HSCRF, word-level labels  are utilized to derive the segment scores.
Further, a CRF output layer and an HSCRF output layer are integrated into a unified neural network and trained jointly.
We evaluate our model on CoNLL 2003 English NER task \cite{tjong2003introduction} and achieve state-of-the-art performance when no external knowledge is used.

In summary, the  contributions of this paper are:
(1) we propose the HSCRF architecture which employs both word-level and segment-level labels for segment score calculation.
(2) we propose a joint CRF-HSCRF training framework and a naive joint decoding algorithm for neural sequence labeling.
(3) we achieve state-of-the-art performance in CoNLL 2003 NER shared task.

\section{Methods}


\begin{figure}[!t]
\centering
\includegraphics[width=2.5in]{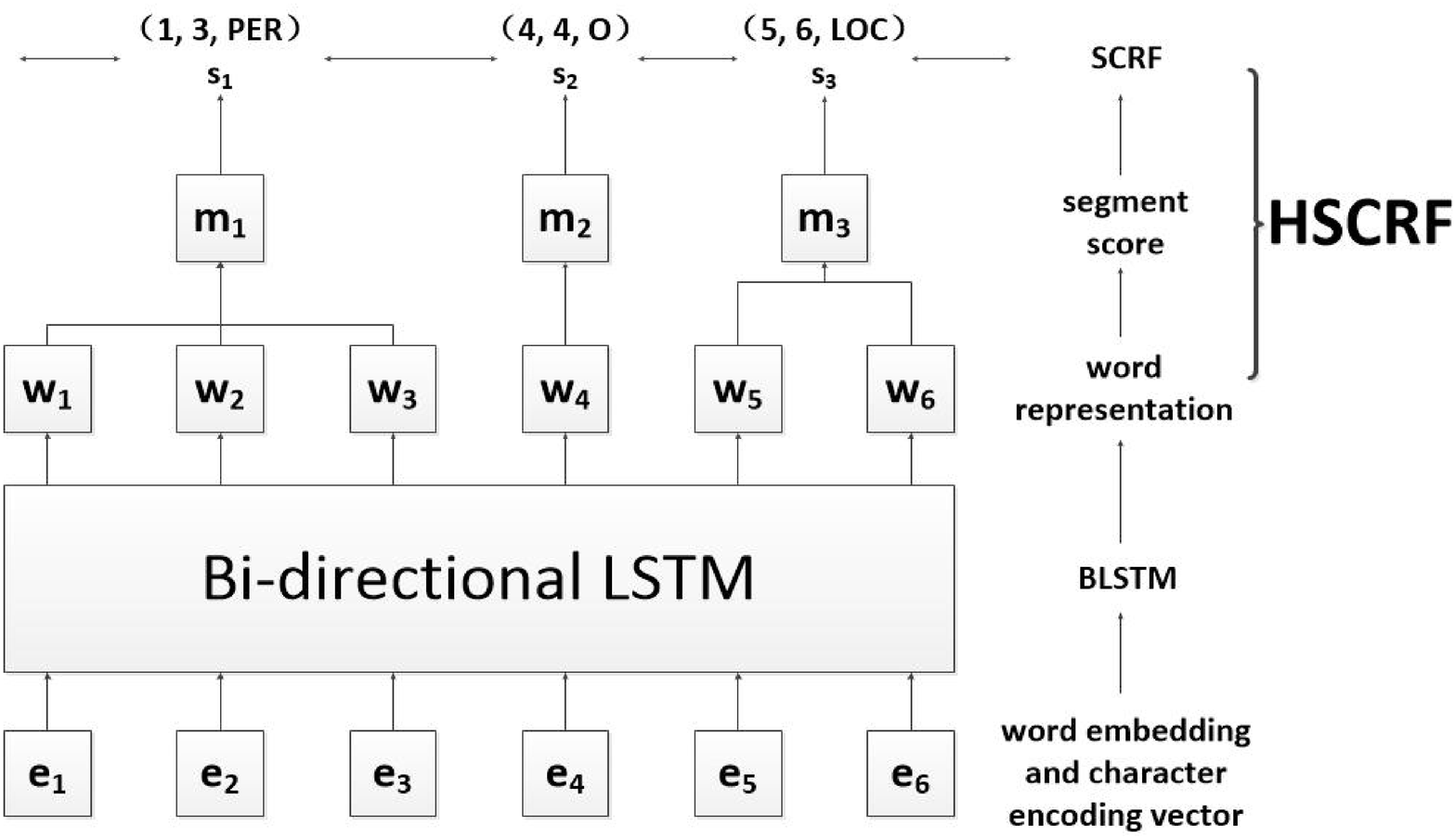}
\caption{The diagram of a neural network with an HSCRF output layer for sequence labeling.}
\label{fig:SCRF}
\end{figure}

\subsection{Hybrid semi-Markov CRFs}
Let $\mathbf{s} = \{ s_{1}, s_{2}, ..., s_{p} \}$ denote the segmentation of an input sentence $\mathbf{x} = \{x_{1}, ..., x_{n}\}$
and $\mathbf{w} = \{w_{1}, ..., w_{n}\}$ denote the sequence of word representations of $\mathbf{x}$ derived by a neural network as shown in Fig. \ref{fig:SCRF}.
Each segment $s_{i} = (b_{i}, e_{i}, l_{i})$, $0 \leq i\leq p$, is a triplet of a begin word index $b_{i}$, an end word index $e_{i}$ and a segment-level label $l_{i}$, where $b_{1} = 1$, $e_{p} = \lvert \mathbf{x} \rvert$, $b_{i+1} = e_{i}+1$, $0 \le e_{i}-b_{i} < L$, and $L$ is the upperbound of the length of $s_{i}$.
Correspondingly, let $\mathbf{y} = \{y_{1}, ..., y_{n}\}$ denote the word-level labels of $\mathbf{x}$.
For example, if a sentence $\mathbf{x}$ in NER task is ``Barack Hussein Obama and Natasha Obama",
we have the corresponding $\mathbf{s}$ = $((1,3,PER), (4,4,O), (5,6,PER))$ and $\mathbf{y}$ = (B-PER, I-PER, E-PER, O, B-PER, E-PER).

Similar to conventional SCRFs \cite{sarawagi2005semi}, the probability of a segmentation $\hat{\mathbf{s}}$ in an HSCRF is defined as
\begin{equation}
\mathrm{p}(\hat{\mathbf{s}}|\mathbf{w}) = \frac{\mathrm{score}(\hat{\mathbf{s}}, \mathbf{w})}{\sum_{\mathbf{s^{'}} \in \mathbf{S}} \mathrm{score}(\mathbf{s^{'}}, \mathbf{w})},
\end{equation}
where $\mathbf{S}$ contains all possible segmentations and
\begin{equation}
\mathrm{score}(\mathbf{s}, \mathbf{w}) = \prod_{i=1}^{\lvert \mathbf{s} \rvert}\mathrm{\psi} ( l_{i-1}, l_{i}, \mathbf{w}, b_{i}, e_{i} ).
\label{equa:SCRFscore}
\end{equation}
Here, $\mathrm{\psi} ( l_{i-1}, l_{i}, \mathbf{w}, b_{i}, e_{i} ) = exp \{ m_{i} + b_{l_{i-1},l_{i}}\}$,
where $m_{i} = \mathrm{\varphi_{h}}(l_{i}, \mathbf{w}, b_{i}, e_{i})$ is the segment score and  $b_{i,j}$ is the segment-level transition parameter from  class $i$ to class $j$.

Different from existing methods of utilizing SCRFs in neural sequence labeling \cite{zhuo2016segment, kong2015segmental} , the segment score in an HSCRF is calculated using word-level labels as
\begin{equation}
\label{a}
  m_{i} 
          = \sum_{k=b_{i}}^{e_{i}} \mathrm{\varphi_{c}}(y_{k}, \mathbf{w'}_{k})
          = \sum_{k=b_{i}}^{e_{i}} \mathbf{a}^{\top}_{y_{k}}\mathbf{w'}_{k},
\end{equation}
where $\mathbf{w'}_{k}$ is the feature vector of the $k$-th word,
$\mathrm{\varphi_{c}}(y_{k}, \mathbf{w'}_{k})$ calculates the score of the $k$-th word being classified into word-level class $y_{k}$,
and $\mathbf{a}_{y_{k}}$ is a weight parameter vector corresponding to class $y_{k}$.
For each word, $\mathbf{w'}_{k}$ is composed of word representation $\mathbf{w}_{k}$ and another two segment-level descriptions, i.e.,
(1) $\mathbf{w}_{e_{i}}-\mathbf{w}_{b_{i}}$ which is derived based on the assumption that word representations in the same segment (e.g., ``Barack Obama") are closer to each other than otherwise (e.g., ``Obama is"),
and (2) $\mathrm{\bm{\phi}}  (k-b_{i}+1)$ which is the embedding vector of the word index in a segment.
Finally, we have $\mathbf{w}'_{k} = [\mathbf{w}_{k}; \mathbf{w}_{e_{i}} - \mathbf{w}_{b_{i}}; \mathrm{\bm{\phi}} (k-b_{i}+1)]$, where $b_{i} \le k \le e_{i}$ and $[ ; ; ]$ is a vector concatenation operation.

The training and decoding criteria of conventional SCRFs \cite{sarawagi2005semi} are followed.
The negative log-likelihood (NLL), i.e., $-log \text{p} (\hat{\mathbf{s}}|\mathbf{w})$, is minimized to estimate the parameters of the HSCRF layer and the lower neural network layers that derive word representations.
For decoding, the Viterbi algorithm is employed to obtain the optimal segmentation as
\begin{align}
\mathbf{s^{*}} = \underbrace{ \mathrm{argmax} }_{ \mathbf{s'} \in \mathbf{S} } log \text{p}(\mathbf{s'}|\mathbf{m}),
\label{equa:SCRFdecode}
\end{align}
where $\mathbf{S}$ contains all legitimate segmentations.

\subsection{Jointly training and decoding using CRFs and HSCRFs}
To further investigate the effects of word-level labels on the training of SCRFs,
we integrate a CRF output layer and a HSCRF output layer into an unified neural network and train them jointly.
These two output layers share the same sequence of word representations $\mathbf{w}$ which are extracted by lower neural network layers.
Given both word-level and segment-level ground truth labels of training sentences,
the model parameters are optimized by minimizing the summation of the loss functions of the CRF layer and the HSCRF layer  with equal weights.

At decoding time, two label sequences, i.e., $\mathbf{s}_{c}$ and $\mathbf{s}_{h}$, for an input sentence can be obtained using the CRF output layer and the HSCRF output layer respectively.
A naive joint decoding algorithm is also designed to make a selection between them.
Assume the NLLs of measuring $\mathbf{s}_{c}$ and $\mathbf{s}_{h}$ using the CRF and HSCRF layers are $NLL_{c}$ and $NLL_{h}$ respectively.
Then, we exchange the models and measure the NLLs of $\mathbf{s}_{c}$ and $\mathbf{s}_{h}$ by HSCRF and CRF and obtain another two values $NLL_{c\_by\_h}$ and $NLL_{h\_by\_c}$.
We just naively assign the summation of $NLL_{c}$ and $NLL_{c\_by\_h}$ to $\mathbf{s}_{c}$, and the summation of $NLL_{h}$ and $NLL_{h\_by\_c}$  to $\mathbf{s}_{h}$.
Finally, we choose the one between $\mathbf{s}_{c}$ and $\mathbf{s}_{h}$ with lower NLL sum as the final result.

\begin{table*}[t!]
\small
\begin{center}
\begin{tabular}{|c|c|c|c|c|c|c|}
\hline
\bf No. & \bf Model Name       &  \bf Word Representation & \bf Top Layer & \bf Decoding Layer  & \bf F1 Score ($\pm $std)  \\ \hline
1  &CNN-BLSTM-CRF       &  CNN-BLSTM       & CRF     &  CRF    & $90.92\pm 0.08$ \\ \hline
2  &CNN-BLSTM-GSCRF     &  CNN-BLSTM       & GSCRF   &  GSCRF  & $90.96\pm 0.12$ \\ \hline
3  &CNN-BLSTM-HSCRF     &  CNN-BLSTM       & HSCRF   &  HSCRF  & $91.10\pm 0.12$ \\ \hline
4  &CNN-BLSTM-JNT(CRF)  &  CNN-BLSTM       & CRF+HSCRF &  CRF    & $91.08\pm 0.12$ \\ \hline
5  &CNN-BLSTM-JNT(HSCRF)&  CNN-BLSTM       & CRF+HSCRF &  HSCRF  & $91.20\pm 0.10$ \\ \hline
6  &CNN-BLSTM-JNT(JNT)  & CNN-BLSTM        & CRF+HSCRF &  CRF+HSCRF& $91.26\pm 0.10$ \\ \hline
7  &LM-BLSTM-CRF        &  LM-BLSTM  & CRF     &  CRF    & $91.17\pm 0.11$ \\ \hline
8  &LM-BLSTM-GSCRF      &  LM-BLSTM  & GSCRF   &  GSCRF  & $91.06\pm 0.05$ \\ \hline
9  &LM-BLSTM-HSCRF      &  LM-BLSTM  & HSCRF   &  HSCRF  & $91.27\pm 0.08$ \\ \hline
10 &LM-BLSTM-JNT(CRF)   &  LM-BLSTM  & CRF+HSCRF &  CRF    & $91.24\pm 0.07$ \\ \hline
11 &LM-BLSTM-JNT(HSCRF) &  LM-BLSTM  & CRF+HSCRF &  HSCRF  & $91.34\pm 0.10$ \\ \hline
12 &LM-BLSTM-JNT(JNT)   &  LM-BLSTM  & CRF+HSCRF &  CRF+HSCRF& $91.38\pm 0.10$ \\ \hline
\end{tabular}
\end{center}
\caption{\label{Tab:allmodels} Model descriptions and their performance on CoNLL 2003 NER task. }
\end{table*}

\section{Experiments}

\subsection{Dataset}
We evaluated our model on the CoNLL 2003 English NER dataset \cite{tjong2003introduction}.
This dataset contained four labels of named entities (PER, LOC, ORG and MISC) and label O for others.
The existing separation of training, development and test sets was followed in our experiments.
We adopted the same word-level tagging scheme as the one used in \citet{liu2017empower} (e.g., BIOES instead of BIO). 
For better computation efficiency, the max segment length $L$ introduced in Section 2.1 was set to 6,
which pruned less than $0.5\%$ training sentences for building SCRFs and had no effect on the development and test sets.

\subsection{Implementation}
As shown in Fig. \ref{fig:SCRF}, the  GloVe \cite{pennington2014glove} 
word embedding and the character encoding vector of each word in the input sentence were concatenated
and fed into a bi-directional LSTM to obtain the sequence of word representations $\mathbf{w}$.
Two character encoding models, LM-BLSTM \cite{liu2017empower} and CNN-BLSTM \cite{ma2016end}, were adopted in our experiments.
Regarding with the top classification layer, we compared our proposed HSCRF with conventional word-level CRF and
grSemi-CRF (GSCRF) \cite{zhuo2016segment}, which was an SCRF using only segment-level information.
The descriptions of the models built in our experiments are summarized in Table \ref{Tab:allmodels}.

For a fair comparison, we implemented all models in the same framework using PyTorch library\footnote{\url{http://pytorch.org/}}.
The hyper-parameters of the models are shown in Table \ref{Tab:hyper} and they were selected according to the two baseline methods without fine-tuning.
Each model in Table \ref{Tab:allmodels} was estimated 10 times and its mean and standard deviation of F1 score were reported
considering the influence of randomness and the weak correlation between development set and test set in this task \cite{Nils2017Reporting}.

\begin{table}[t!]
\small
\begin{center}
\begin{tabular}{|p{1.1in}<{\centering}|p{0.7in}<{\centering}|p{0.6in}<{\centering}|}

\hline
\bf Component                                       & \bf Parameter & \bf Value\\ \hline
word-level & {\multirow{2}{*}{dimension}} & {\multirow{2}{*}{100}} \\
embedding$^{\dagger \ddagger}$ && \\ \hline
character-level & {\multirow{2}{*}{dimension}} & {\multirow{2}{*}{30}} \\
embedding$^{\dagger \ddagger}$ && \\ \hline
{\multirow{2}{*}{character-level LSTM$^{\dagger}$}} & depth         & 1        \\ \cline{2-3}
                                                    & hidden size   & 300      \\ \hline
highway network$^{\dagger}$                         & layer         & 1        \\ \hline

{\multirow{2}{*}{word-level BLSTM$^{\dagger}$}}     & depth         & 1        \\ \cline{2-3}
                                                    & hidden size   & 300      \\ \hline
{\multirow{2}{*}{word-level BLSTM$^{\ddagger}$}}    & depth         & 1        \\ \cline{2-3}
                                                    & hidden size   & 200      \\ \hline

{\multirow{2}{*}{CNN$^{\ddagger}$}}                 & window size   & 3        \\ \cline{2-3}
                                                    & filter number & 30       \\ \hline
$\bm{\phi}(\cdot)^{\dagger \ddagger}$              & dimension     & 10       \\ \hline
dropout$^{\dagger \ddagger}$                        & dropout rate  & 0.5      \\ \hline
{\multirow{5}{*}{optimization$^{\dagger \ddagger}$}}& learning rate & 0.01     \\ \cline{2-3}
                                                    & batch size    & 10       \\ \cline{2-3}
                                                    & strategy      & SGD      \\ \cline{2-3}
                                                    & gradient clip & 5.0      \\   \cline{2-3}
                                                    & decay rate    & 1/(1+0.05t) \\ \hline
\end{tabular}
\end{center}
\caption{\label{Tab:hyper}Hyper-parameters of the models built in our experiments, where $\dagger$ indicates the ones when using LM-BLSTM for deriving word representations and
$\ddagger$ indicates the ones when using CNN-BLSTM.}
\end{table}

\subsection{Results}
Table \ref{Tab:allmodels} lists the F1 score results of all built models on CoNLL 2003 NER task.
Comparing model 3 with model 1/2 and model 9 with model 7/8,  we can see that HSCRF performed better than CRF and GSCRF.
The superiorities were significant since the $p$-values of $t$-test were smaller than $0.01$. 
This implies the benefits of utilizing word-level labels when deriving segment scores in SCRFs.
Comparing model 1 with model 4, 3 with 5, 7 with 10, and 9 with 11, 
we can see that the jointly training method introduced in Section 2.2 improved the performance of CRF and HSCRF significantly ($p < 0.01$ in all these four pairs).
This may be attributed to that jointly training generates better word representations that can be shared by both CRF and HSCRF decoding layers.
Finally, comparing model 6 with model 4/5 and model 12 with model 10/11, we can see the effectiveness of the jointly decoding algorithm introduced in Section 2.2
on improving F1 scores ($p < 0.01$ in all these four pairs).
The LM-BLSTM-JNT model with jointly decoding achieved the highest F1 score among all these built models.

\begin{table}[t!]
\small
\begin{center}
\begin{tabular}{|c|c|c|}
\hline
{\multirow{2}{*}{\bf Model}}   & \multicolumn{2}{c|}{\bf Test Set F1 Score} \\ \cline{2-3}
                                                        & \bf Type   & \bf Value ($\pm $std)\\
\hline
\citet{zhuo2016segment}                                 & reported   & 88.12              \\ \hline
\citet{lample2016neural}                                & reported   & 90.94              \\ \hline
\citet{ma2016end}                                       & reported   & 91.21              \\ \hline
\citet{rei2017semi}                      & reported   & 86.26              \\ \hline
{\multirow{2}{*}{\citet{liu2017empower}}}
                                                        & mean       & 91.24 $\pm$ 0.12
\\ \cline{2-3}
                                                        & max        & 91.35
\\ \hline
{\multirow{2}{*}{CNN-BLSTM-CRF}}                        & mean       & 90.92 $\pm$ 0.08    \\ \cline{2-3}
                                                        & max        & 91.04
\\ \hline
{\multirow{2}{*}{LM-BLSTM-CRF}}           & mean       & 91.17 $\pm$ 0.11    \\ \cline{2-3}
                                                        & max        & 91.30
\\ \hline
{\multirow{2}{*}{CNN-BLSTM-JNT(JNT)}}                   & mean        & 91.26 $\pm$ 0.10 \\ \cline{2-3}
                                                        & max         & 91.41           \\
\hline
{\multirow{2}{*}{LM-BLSTM-JNT(JNT)}}      & mean        & \textbf{91.38}$\pm$ \textbf{0.10}     \\ \cline{2-3}
                                                        & max         & \textbf{91.53}
\\ \hline
\hline
\citet{luo2015joint}$^*$                       & reported   & 91.2               \\ \hline
\citet{chiu2015named}$^*$                      & reported   & 91.62 $\pm$ 0.33    \\ \hline
\citet{tran2017named}$^*$              & reported   & 91.66              \\ \hline
\citet{peters2017semi}$^*$             & reported   & 91.93 $\pm$ 0.19    \\ \hline
\citet{yang2017transfer}$^*$                   & reported   & 91.26              \\ \hline
\end{tabular}
\end{center}
\caption{\label{Tab:allNERresults} Comparison with existing work on CoNLL 2003 NER task. The models labelled with $^*$ utilized external knowledge beside CoNLL 2003 training set and pre-trained word embeddings. }
\end{table}

\begin{table*}[t!]
\small
\begin{center}
\begin{tabular}{|c|c|c|c|c|c|c|c|c|}
\hline

{\multirow{2}{*}{\bf No.}}&{\multirow{2}{*}{\bf Model Name}}   & \multicolumn{7}{c|}{\bf Entity Length} \\ \cline{3-9}
&& \bf 1  &  \bf 2 & \bf 3 & \bf 4  & \bf 5 & \bf $\mathbf{\geq}$ 6 & \bf all  \\ \hline

7&LM-BLSTM-CRF      & 91.68        & 91.88        &82.64         & 75.81         & 73.68         & 72.73             & 91.17        \\ \hline
8&LM-BLSTM-GSCRF    & 91.57        & 91.68        &83.61         & 74.32         & 76.64            & 73.64             & 91.06        \\ \hline
9&LM-BLSTM-HSCRF    & 91.65        & 91.84        &82.97         & 76.20         & 78.95    &74.55              & 91.27        \\ \hline
12&LM-BLSTM-JNT(JNT)&\textbf{91.73}&\textbf{92.03}&\textbf{83.78}&\textbf{77.27} &\textbf{79.66} &\textbf{76.55}     &\textbf{91.38}\\ \hline
\end{tabular}
\end{center}
\caption{\label{Tab:Rresultlength}Model performance on CoNLL 2003 NER task for entities with different lengths.}
\end{table*}

\subsection{Comparison with existing work}
Table \ref{Tab:allNERresults} shows some recent results\footnote{It should be noticed that the results of \citet{liu2017empower} were inconsistent with
the original ones reported in their paper. According to its first author's GitHub page (https://github.com/LiyuanLucasLiu/LM-LSTM-CRF), the originally reported results had errors due to some bugs. Here, we report the results after the bugs got fixed.} on the CoNLL 2003 English NER task.
For the convenience of comparison, we also listed the maximum F1 scores among 10 repetitions when building our models.
The maximum F1 score of our re-implemented CNN-BLSTM-CRF model was slightly worse than the one originally reported in \citet{ma2016end},
but it was similar to the one reported in \citet{Nils2017Reporting}.


In the NER models listed in Table \ref{Tab:allNERresults}, \citet{zhuo2016segment} employed some manual features and calculated segment scores by grConv for SCRF. \citet{lample2016neural} and \citet{ma2016end} constructed character-level encodings using BLSTM and CNN respectively, and concatenated them with word embeddings.
Then, the same BLSTM-CRF architecture was adopted in both models.
\citet{rei2017semi} fed word embeddings into LSTM to obtain the word representations for CRF decoding and to predict the next word simultaneously.
Similarly, \citet{liu2017empower} input characters into LSTM to predict the next character and to get the character-level encoding for each word.

Some of the models listed in Table \ref{Tab:allNERresults} utilized external knowledge beside CoNLL 2003 training set and pre-trained word embeddings.
\citet{luo2015joint} proposed JERL model, which was trained on both NER and entity linking tasks simultaneously.
\citet{chiu2015named} employed lexicon features from DBpedia \cite{auer2007dbpedia}.
\citet{tran2017named} and \citet{peters2017semi} utilized pre-trained language models from large corpus to model word representations.
\citet{yang2017transfer} utilized transfer learning to obtain shared information from other tasks, such as chunking and POS tagging, for word representations.



From Table \ref{Tab:allNERresults}, we can see that our CNN-BLSTM-JNT and LM-BLSTM-JNT models with jointly decoding both achieved state-of-the-art
F1 scores among all models without using external knowledge.
The maximum F1 score achieved by the  LM-BLSTM-JNT model was $91.53\%$.

\subsection{Analysis}

To better understand the effectiveness of word-level and segment-level labels on the NER task, we evaluated the performance of  models 7, 8, 9 and 12
in Table \ref{Tab:allNERresults} for entities with different lengths.
The mean F1 scores of 10 training repetitions are reported in Table \ref{Tab:Rresultlength}.
Comparing model 7 with model 8, we can see that GSCRF achieved better performance than CRF for long entities (with more than 4 words) but worse for short entities (with less than 3 words).
Comparing model 7 with model 9, we can find that HSCRF outperformed CRF for recognizing long entities and meanwhile achieved comparable performance with CRF for short entities.

One possible explanation is that word-level labels may supervise models to learn word-level descriptions which tend to benefit the recognition of short entities.
On the other hand, segment-level labels may guide models to capture the descriptions of combining words for whole entities which help to recognize long entities.
By utilizing both labels, the LM-BLSTM-HSCRF model can achieve better overall performance of recognizing entities with different lengths.
Furthermore, the LM-BLSTM-JNT(JNT) model which adopted jointly training and decoding
achieved the best performance among all models shown in Table \ref{Tab:Rresultlength} for all entity lengths.

\section{Conclusions}
This paper proposes a hybrid semi-Markov conditional random field (HSCRF) architecture for neural sequence labeling,
in which word-level labels are utilized to derive the segment scores in SCRFs.
Further, the methods of training and decoding CRF and HSCRF output layers jointly are also presented.
Experimental results on CoNLL 2003 English NER task demonstrated the effectiveness of the proposed HSCRF model which achieved state-of-the-art performance.

\bibliography{acl2018}

\begin{thebibliography}{20}
\expandafter\ifx\csname natexlab\endcsname\relax\def\natexlab#1{#1}\fi

\bibitem[{Andrew(2006)}]{andrew2006hybrid}
Galen Andrew. 2006.
\newblock \href {http://www.aclweb.org/anthology/W06-1655} {A hybrid
  {M}arkov/semi-{M}arkov conditional random field for sequence segmentation}.
\newblock In \emph{Proceedings of the 2006 Conference on Empirical Methods in
  Natural Language Processing}, pages 465--472. Association for Computational
  Linguistics.

\bibitem[{Auer et~al.(2007)Auer, Bizer, Kobilarov, Lehmann, Cyganiak, and
  Ives}]{auer2007dbpedia}
S{\"o}ren Auer, Christian Bizer, Georgi Kobilarov, Jens Lehmann, Richard
  Cyganiak, and Zachary Ives. 2007.
\newblock Dbpedia: A nucleus for a web of open data.
\newblock In \emph{The semantic web}, pages 722--735. Springer.

\bibitem[{Chiu and Nichols(2016)}]{chiu2015named}
Jason Chiu and Eric Nichols. 2016.
\newblock \href {http://www.aclweb.org/anthology/Q16-1026} {Named entity
  recognition with bidirectional {LSTM-CNNs}}.
\newblock \emph{Transactions of the Association of Computational Linguistics},
  4:357--370.

\bibitem[{Huang et~al.(2015)Huang, Xu, and Yu}]{huang2015bidirectional}
Zhiheng Huang, Wei Xu, and Kai Yu. 2015.
\newblock Bidirectional {LSTM-CRF} models for sequence tagging.
\newblock \emph{arXiv preprint arXiv:1508.01991}.

\bibitem[{Kong et~al.(2015)Kong, Dyer, and Smith}]{kong2015segmental}
Lingpeng Kong, Chris Dyer, and Noah~A Smith. 2015.
\newblock Segmental recurrent neural networks.
\newblock \emph{arXiv preprint arXiv:1511.06018}.

\bibitem[{Lafferty et~al.(2001)Lafferty, McCallum, and
  Pereira}]{lafferty2001conditional}
John Lafferty, Andrew McCallum, and Fernando~CN Pereira. 2001.
\newblock Conditional random fields: Probabilistic models for segmenting and
  labeling sequence data.

\bibitem[{Lample et~al.(2016)Lample, Ballesteros, Subramanian, Kawakami, and
  Dyer}]{lample2016neural}
Guillaume Lample, Miguel Ballesteros, Sandeep Subramanian, Kazuya Kawakami, and
  Chris Dyer. 2016.
\newblock \href {https://doi.org/10.18653/v1/N16-1030} {Neural architectures
  for named entity recognition}.
\newblock In \emph{Proceedings of the 2016 Conference of the North American
  Chapter of the Association for Computational Linguistics: Human Language
  Technologies}, pages 260--270. Association for Computational Linguistics.

\bibitem[{{Liu} et~al.(2018){Liu}, {Shang}, {Xu}, {Ren}, {Gui}, {Peng}, and
  {Han}}]{liu2017empower}
L.~{Liu}, J.~{Shang}, F.~{Xu}, X.~{Ren}, H.~{Gui}, J.~{Peng}, and J.~{Han}.
  2018.
\newblock {Empower Sequence Labeling with Task-Aware Neural Language Model}.
\newblock In \emph{AAAI}.

\bibitem[{Luo et~al.(2015)Luo, Huang, Lin, and Nie}]{luo2015joint}
Gang Luo, Xiaojiang Huang, Chin-Yew Lin, and Zaiqing Nie. 2015.
\newblock \href {https://doi.org/10.18653/v1/D15-1104} {Joint entity
  recognition and disambiguation}.
\newblock In \emph{Proceedings of the 2015 Conference on Empirical Methods in
  Natural Language Processing}, pages 879--888. Association for Computational
  Linguistics.

\bibitem[{Ma and Hovy(2016)}]{ma2016end}
Xuezhe Ma and Eduard Hovy. 2016.
\newblock \href {https://doi.org/10.18653/v1/P16-1101} {End-to-end sequence
  labeling via bi-directional {LSTM-CNNs-CRF}}.
\newblock In \emph{Proceedings of the 54th Annual Meeting of the Association
  for Computational Linguistics (Volume 1: Long Papers)}, pages 1064--1074.
  Association for Computational Linguistics.

\bibitem[{Mikolov et~al.(2013)Mikolov, Sutskever, Chen, Corrado, and
  Dean}]{mikolov2013distributed}
Tomas Mikolov, Ilya Sutskever, Kai Chen, Greg~S Corrado, and Jeff Dean. 2013.
\newblock Distributed representations of words and phrases and their
  compositionality.
\newblock In \emph{Advances in neural information processing systems}, pages
  3111--3119.

\bibitem[{Pennington et~al.(2014)Pennington, Socher, and
  Manning}]{pennington2014glove}
Jeffrey Pennington, Richard Socher, and Christopher Manning. 2014.
\newblock \href {https://doi.org/10.3115/v1/D14-1162} {Glo{V}e: Global vectors
  for word representation}.
\newblock In \emph{Proceedings of the 2014 Conference on Empirical Methods in
  Natural Language Processing (EMNLP)}, pages 1532--1543. Association for
  Computational Linguistics.

\bibitem[{Peters et~al.(2017)Peters, Ammar, Bhagavatula, and
  Power}]{peters2017semi}
Matthew Peters, Waleed Ammar, Chandra Bhagavatula, and Russell Power. 2017.
\newblock \href {https://doi.org/10.18653/v1/P17-1161} {Semi-supervised
  sequence tagging with bidirectional language models}.
\newblock In \emph{Proceedings of the 55th Annual Meeting of the Association
  for Computational Linguistics (Volume 1: Long Papers)}, pages 1756--1765.
  Association for Computational Linguistics.

\bibitem[{Rei(2017)}]{rei2017semi}
Marek Rei. 2017.
\newblock \href {https://doi.org/10.18653/v1/P17-1194} {Semi-supervised
  multitask learning for sequence labeling}.
\newblock In \emph{Proceedings of the 55th Annual Meeting of the Association
  for Computational Linguistics (Volume 1: Long Papers)}, pages 2121--2130.
  Association for Computational Linguistics.

\bibitem[{Reimers and Gurevych(2017)}]{Nils2017Reporting}
Nils Reimers and Iryna Gurevych. 2017.
\newblock \href {http://aclweb.org/anthology/D17-1035} {Reporting score
  distributions makes a difference: Performance study of {LSTM}-networks for
  sequence tagging}.
\newblock In \emph{Proceedings of the 2017 Conference on Empirical Methods in
  Natural Language Processing}, pages 338--348. Association for Computational
  Linguistics.

\bibitem[{Sang and Meulder(2003)}]{tjong2003introduction}
Erik F. Tjong~Kim Sang and Fien~De Meulder. 2003.
\newblock \href {http://www.aclweb.org/anthology/W03-0419} {Introduction to the
  {CoNLL}-2003 shared task: Language-independent named entity recognition}.
\newblock In \emph{Proceedings of the Seventh Conference on Natural Language
  Learning at HLT-NAACL 2003}.

\bibitem[{Sarawagi and Cohen(2005)}]{sarawagi2005semi}
Sunita Sarawagi and William~W Cohen. 2005.
\newblock Semi-{M}arkov conditional random fields for information extraction.
\newblock In \emph{Advances in neural information processing systems}, pages
  1185--1192.

\bibitem[{Tran et~al.(2017)Tran, MacKinlay, and Jimeno~Yepes}]{tran2017named}
Quan Tran, Andrew MacKinlay, and Antonio Jimeno~Yepes. 2017.
\newblock \href {http://aclweb.org/anthology/I17-1057} {Named entity
  recognition with stack residual {LSTM} and trainable bias decoding}.
\newblock In \emph{Proceedings of the Eighth International Joint Conference on
  Natural Language Processing (Volume 1: Long Papers)}, pages 566--575. Asian
  Federation of Natural Language Processing.

\bibitem[{Yang et~al.(2017)Yang, Salakhutdinov, and Cohen}]{yang2017transfer}
Zhilin Yang, Ruslan Salakhutdinov, and William~W Cohen. 2017.
\newblock Transfer learning for sequence tagging with hierarchical recurrent
  networks.
\newblock \emph{arXiv preprint arXiv:1703.06345}.

\bibitem[{Zhuo et~al.(2016)Zhuo, Cao, Zhu, Zhang, and Nie}]{zhuo2016segment}
Jingwei Zhuo, Yong Cao, Jun Zhu, Bo~Zhang, and Zaiqing Nie. 2016.
\newblock \href {https://doi.org/10.18653/v1/P16-1134} {Segment-level sequence
  modeling using gated recursive semi-{M}arkov conditional random fields}.
\newblock In \emph{Proceedings of the 54th Annual Meeting of the Association
  for Computational Linguistics (Volume 1: Long Papers)}, pages 1413--1423.
  Association for Computational Linguistics.

\end{thebibliography}
\bibliographystyle{acl_natbib}

\end{document}